\documentclass[a4paper, 12 pt, onecolumn]{ieeeconf}  

\IEEEoverridecommandlockouts                              

\usepackage[cmex10]{amsmath, mathtools}
\usepackage{amsfonts, amssymb, amsthm}
\usepackage{algorithmic}
\usepackage{algorithm}
\usepackage{array}
\usepackage[usenames,dvipsnames]{color}
\usepackage[caption=false,font=footnotesize]{subfig}
\usepackage{setspace}
\usepackage{cite}

\theoremstyle{plain}

\theoremstyle{definition}

\newcommand{\note}{\color{red}\bf}

\title{\LARGE \bf
Optimal Control-Based UAV Path Planning with Dynamically-Constrained TSP with Neighborhoods
}

\author{Dae-Sung Jang$^{1}$, Hyeok-Joo Chae$^{1}$, and Han-Lim Choi$^{1}$
\thanks{$^{1}$The authors are with Department of Aerospace Engineering, KAIST, Daejeon 34141, Republic of Korea.
	E-mail: {\tt\small \{dsjang, hjchae\}@lics.kaist.ac.kr}, {\tt\small hanlimc@kaist.ac.kr}.}%
}

\begin{document}

\setstretch{1.5}

\maketitle
\thispagestyle{empty}
\pagestyle{empty}

\begin{abstract}
	
\iftrue
This paper addresses path planning of an unmanned aerial vehicle (UAV) with remote sensing capabilities (or wireless communication capabilities).
The goal of the path planning is to find a minimum-flight-time closed tour of the UAV visiting all executable areas of given remote sensing and communication tasks; in order to incorporate the nonlinear vehicle dynamics, this problem is regarded as a dynamically-constrained traveling salesman problem with neighborhoods.
To obtain a close-to-optimal solution for the path planning in a tractable manner, a sampling-based roadmap algorithm that embeds an optimal control-based path generation process is proposed.
The algorithm improves the computational efficiency by reducing numerical computations required for optimizing inefficient local paths, and by extracting additional information from a roadmap of a fixed number of samples.
Comparative numerical simulations validate the efficiency of the presented algorithm in reducing computation time and improving the solution quality compared to previous roadmap-based planning methods.
\fi

\end{abstract}

\section{Introduction}

Path planning is one of the key problems on automated operation of unmanned aerial vehicles (UAVs) in reducing costs, flight time, fuel consumption, and chemical emission.
A traditional technique on UAV operation is that a UAV equipped with attitude, heading, and speed controllers under a guidance rule navigates and tracks certain waypoints in a given order of visiting specified by a human operator.
In a higher level of the UAV automation scheme, called decision-making, even the visiting order of given tasks is determined by the UAV itself in accordance with its mission objective.
In the case that the tasks are assigned point-wise and performed by visiting the exact locations, the UAV path planning with decision-making has been modeled as the traveling salesman problem (TSP) for a Dubins vehicle \cite{Sal08,Ny12}.

This paper focuses a generalized path planning problem of a UAV having extended capabilities of processing tasks:
such as remote sensing capabilities for surveillance (e.g. electro-optical imaging, synthetic aperture radar, and lidar based ranging/imaging) or wireless communication capabilities for collecting data from a sensor network with low-energy transmitters;
a similar issue to the latter case has been raised in sensor network studies as {\it data mule} problems \cite{Sha03,Jea05,Yua07,Fra11,Liu13,He13,Cha15}.
A UAV with such capabilities does not need to visit the exact position for processing each task, but rather it can just pass through certain region around the task point specified by remote sensing/communication ranges.
The decision-making level of the path planning with a UAV of this type can be represented as the TSP with neighborhoods (TSPN) \cite{Ark94,Dum03,Ber05,Beh14}, where the executable regions of tasks are defined as neighborhoods of nodes.

\iffalse
The TSPN formulation is suitable for vehicles of very high mobility, whose point-to-point visits are approximated by straight line segments.
However, in some applications with comparatively small neighborhoods of flocked tasks, where non-holonomic constraints of a UAV become dominant factors on the flight path, numerical integration might be required for a precise path generation.
\else
However, the TSPN formulation is only appropriate for vehicles of very high mobility, whose point-to-point visits are approximated by straight line segments.
\fi
With an assumption on a UAV that the flight path curvature is constrained in a fixed speed, the path planning problem has been formulated as the Dubins TSPN (DTSPN).
There have been several studies on this topic using evolutionary computations \cite{Obe09,Mac12,Xin14} in a global optimization framework and {\it sampling-based roadmap} methods \cite{Obe12,Isa13}.
A sampling-based roadmap method is a discretization approach using a finite number of state samples distributed in the space, and the roadmap is composed of the optimal paths between the samples.
The roadmap discretizes the problem into the generalized TSP (GTSP) and it is solved by a TSP solver after the Noon-Bean transfomation \cite{Noo93}.
In \cite{Obe12}, the implementations and the resolution completeness of the sampling-based roadmap was analyzed to solve DTSPN with polygonal neighborhoods.
In a roadmap, some sample on a neighborhood can also be contained in other intersecting neighborhoods, which means that a path though the sample visits two tasks' neighborhoods at the same time.
The intersecting neighborhoods of the samples of this kind were considered in \cite{Isa13} for constructing a more informative roadmap, and it is shown that the method in \cite{Isa13} is highly advantageous for densely distributed neighborhoods and theoretically performs better than the method in \cite{Obe12}.


With Dubins path assumptions, only the path length or the flight time under a constant speed can be handled as a cost metric in the UAV path planning.
Furthermore, in a two dimensional top down view of UAV path planning, the dynamics of a UAV including variable speed plays a crucial role in determining maneuverability (e.g. turning radius) and costs, especially flight time.
Thus, for an effective and practical planning, instead of the sampling-based roadmap with shortest Dubins paths of a fixed turning radius and a fixed speed, the roadmap composed of optimal paths obeying the UAV dynamics with variable speed can be preferable.

An optimal control approach for constructing the optimal path connecting two state samples can take into account more complex dynamics of UAVs and various types of cost metrics, such as integrated control inputs, fuel consumption, and carbon emissions, which are the functions of the history of states and inputs of the UAVs.
However, the computational burdens for optimal control paths for every pair of samples from different task neighborhoods grow with the square of the number of tasks, and may dominate the total computation time for the path planning.

Therefore, in this paper, three techniques are introduced into a sampling-based roadmap algorithm to reduce the computation for the UAV path planning with nonlinear dynamics.
The techniques are proposed to improve solution qualities with a reduced computation, but sacrificing the theoretical resolution completeness, which are only achievable with a very large number of samples. 
The first technique directly reduces the number of optimal control paths by exempting the calculations of certain paths that are hardly possible in the optimal solution.
The second one extends the concept of intersecting neighborhoods in \cite{Isa13} to {\it necessarily intersecting neighborhoods} on the trajectory of a UAV following certain dynamics, so that the utilization of the sampled roadmap is enhanced in a TSP solver.
Finally, the solution path of the GTSP instance from the roadmap is locally refined by an optimal control method with preserving the visiting order of neighborhoods.
Numerical simulations demonstrate the efficiency of the presented algorithms in both the computational load reduction and the performance improvement compared with other roadmap methods.


\section{Problem Formulation}\label{sec:pf}
This section provides a mathematical formulation for the problem considered throughout this paper: a two dimensional path planning of a UAV in level flight with turns.
The objective of the problem is to minimize the flight time for a closed tour visiting all neighborhoods of given tasks.
The neighborhood $N_i\subset\mathbb{R}^2$ of a task $i$ is assumed to be a disk centered on the location of the task, and the collection of all the neighborhoods is denoted by $\mathcal{N}$.

The vehicle dynamics is modeled as a curvature constrained Dubins vehicle with variable speed in the plane, as in (\ref{eq:dyn}).
In this model, the UAV can select whether to take slow but small radius turning or to fly faster with sacrificing the agility.
\begin{eqnarray}\label{eq:dyn}
	\dot{x} & = & v\cos{\theta}\nonumber\\
	\dot{y} & = & v\sin{\theta}\nonumber\\
	\dot{v} & = & c_1u_1\\
	\dot{\theta} & = & \frac{c_2}{v}u_2\nonumber
\end{eqnarray}
The coordinates of the vehicle position are $x$ and $y$; $v$ is the speed of the vehicle bounded between $v_{\min}$ and $v_{\max}$; and $\theta$ is the heading of it.
This system has two control inputs $u_1$ and $u_2$.
$u_1$ represents excessive thrust per mass providing an acceleration to the vehicle and $c_1$ is a corresponding constant.
In this paper, $u_1$ is normalized by choosing $c_1$ appropriately so that $u_1\in[-1,1]$.
$u_2$ denotes the steering command that ranges between $[-1,1]$; a negative command signifies a left turn and a positive one steers right.
If the vehicle takes coordinate turns in level flight, the constant for steering $c_2$ can be assumed to be $g\sqrt{n_{\max}^2-1}$, where $g$ is the gravitational acceleration and $n_{\max}$ denotes the maximum load factor of the vehicle.

The formulation of the vehicle model above has an ambiguity on the heading angle, and thus may be unstable in numerical processes of optimal control.
The dynamics can be restated as a system of nonlinear second order differential equations for the coordinates:
\begin{eqnarray}\label{eq:dyn2}
	\ddot{x} & = & \frac{c_1\dot{x}}{\sqrt{\dot{x}^2+\dot{y}^2}}u_1 - \frac{c_2\dot{y}}{\sqrt{\dot{x}^2+\dot{y}^2}}u_2,\nonumber\\
	\ddot{y} & = & \frac{c_1\dot{y}}{\sqrt{\dot{x}^2+\dot{y}^2}}u_1 + \frac{c_2\dot{x}}{\sqrt{\dot{x}^2+\dot{y}^2}}u_2,
\end{eqnarray}
which defines the states of the vehicle without any ambiguities. 

The trajectory $(x(t),y(t))$ of the UAV can be obtained by numerically integrating the differential equations (\ref{eq:dyn2}) from some initial state $(x(t_0),y(t_0))$ at $t=t_0$ with a control input sequence $(u_1(s),u_2(s))$ in $s\in[t_0,t]$.
For given $n$ tasks and their neighborhoods in the plane, a solution for the path planning must construct a closed path, i.e. a tour, which passes every neighborhoods at least one time.
Thus, from the initial time $t_0$ to the final flight time $t_f$ completing the tour, there must exist at least one time $t_i$ so that $(x(t_i),y(t_i))\in{N}_i$ for every task $i$, and $(x(t_0),y(t_0))=(x(t_f),y(t_f))$.
Then, the path planning problem is stated as an optimal control problem that is to find a tuple of the initial state $(x(t_0),y(t_0))$ and the control input sequence $(u_1(t),u_2(t))$ in $t\in[t_0,t_f]$ for minimizing $t_f$ such that the trajectory of UAV coordinates satisfies these neighborhood visiting constraints and the differential equations in (\ref{eq:dyn2}).

\section{Solution Approach}\label{sec:alg}

In this section, an algorithm of a sampling-based roadmap approach is presented for solving the UAV path planning with nonlinear dynamics.
The algorithm is briefly summarized in the following subsection with its grounds, followed by the subsections for detailed descriptions of four steps of the algorithm.

\subsection{Algorithm Outline}

In section \ref{sec:pf}, the UAV path planning problem is formulated as an optimal control problem with neighborhood visiting constraints.
However, obtaining a stable optimal solution of a dynamic system with complex path constraints from an one-shot numerical optimization is a hard task in practice.
Thus, a similar framework to the sampling-based roadmap approaches for DTSP and DTSPN is utilized: the path planning problem is approximately solved by a discretization of the continuous state space and a graph transformation to GTSP.
Instead of using Dubins paths as in DTSP and DTSPN, optimal control paths are numerically obtained between quasi-random samples; i.e. the UAV path planning problem is handled as a TSPN for a nonlinear dynamic system: dynamically-constrained TSPN.

The presented algorithm for solving the path planning is partitioned into four steps:
\begin{enumerate}
	\item Given $n$ tasks and their circular neighborhoods, create $m$ samples in the state space of a UAV along the boundary of each neighborhood. Then, construct a reduced roadmap by creating optimally controlled paths between the pairs of the samples, while excluding very curvy paths before numerical calculations.\label{alg:step1} 
	\item Create a GTSP instance for finding the shortest tour in the reduced roadmap, and transform it to ATSP with appropriate modifications for necessarily intersecting neighborhoods.
	\item Using a TSP solver, obtain the optimal (or sub-optimal) solution for the transformed ATSP, and then interpret it to a solution of the GTSP for the reduced roadmap.
	\item With keeping the order of visiting in the GTSP solution, sequentially (or repeatedly) refine the local paths connecting every successive three neighborhoods by using an optimal control solver.
\end{enumerate}

In the problem size under consideration (several dozen tasks), the numerical path generation by an optimal control solver in step \ref{alg:step1} is the most time-consuming process.
With a minimal or no degradation of the final quality of the solution, some devious paths hardly assumed to be in the optimal tour are excluded before the numerical optimal control process.
The shortest Dubins paths are utilized for discriminating the devious paths, and also for initial guessing in the optimal control.

The roadmap constructed with a reduced number of the optimal paths is then represented as a GTSP problem, and converted to an ATSP problem to be solved by a TSP solver.
During the conversion to ATSP, other neighborhoods necessarily intersected in the course of flight through each sample are identified to avoid visiting neighborhoods multiple times (refer \ref{sec:alg_step2} for the detail).
Finally, the path from the optimal visiting sequence in the reduced roadmap is locally refined.
By this procedure, the number of samples for constructing the roadmap needs not to be very large, since a local optimization in the continuous state space is conducted preserving the visiting sequence from the roadmap.

\subsection{Roadmap Construction with Reduced Paths \& Optimal Control}\label{sec:alg_step1}

The sampling-based roadmap for the UAV path planning is constructed with a reduced number of path calculations.
First, a fixed number of samples are generated for the neighborhood of each task in the state space composed of the position, heading, and speed of the UAV for convenience.
Though the samples can be randomly created, but the quasi-random Halton sequence is utilized for more evenly distributed samples in the 4-dimensional state space under following constraints:
the position of the state samples are on only the boundary of the circular neighborhoods; the heading is assigned to be a direction that the UAV enters the neighborhoods; and the speed is selected between $v_{\min}$ and $v_{\max}$. 
Afterwards, the state samples are converted to the positions $(x,y)$ and the velocities $(\dot{x},\dot{y})$ in Cartesian coordinates.

\begin{figure}[t]
	\centerline{
		\includegraphics[width=.65\columnwidth]{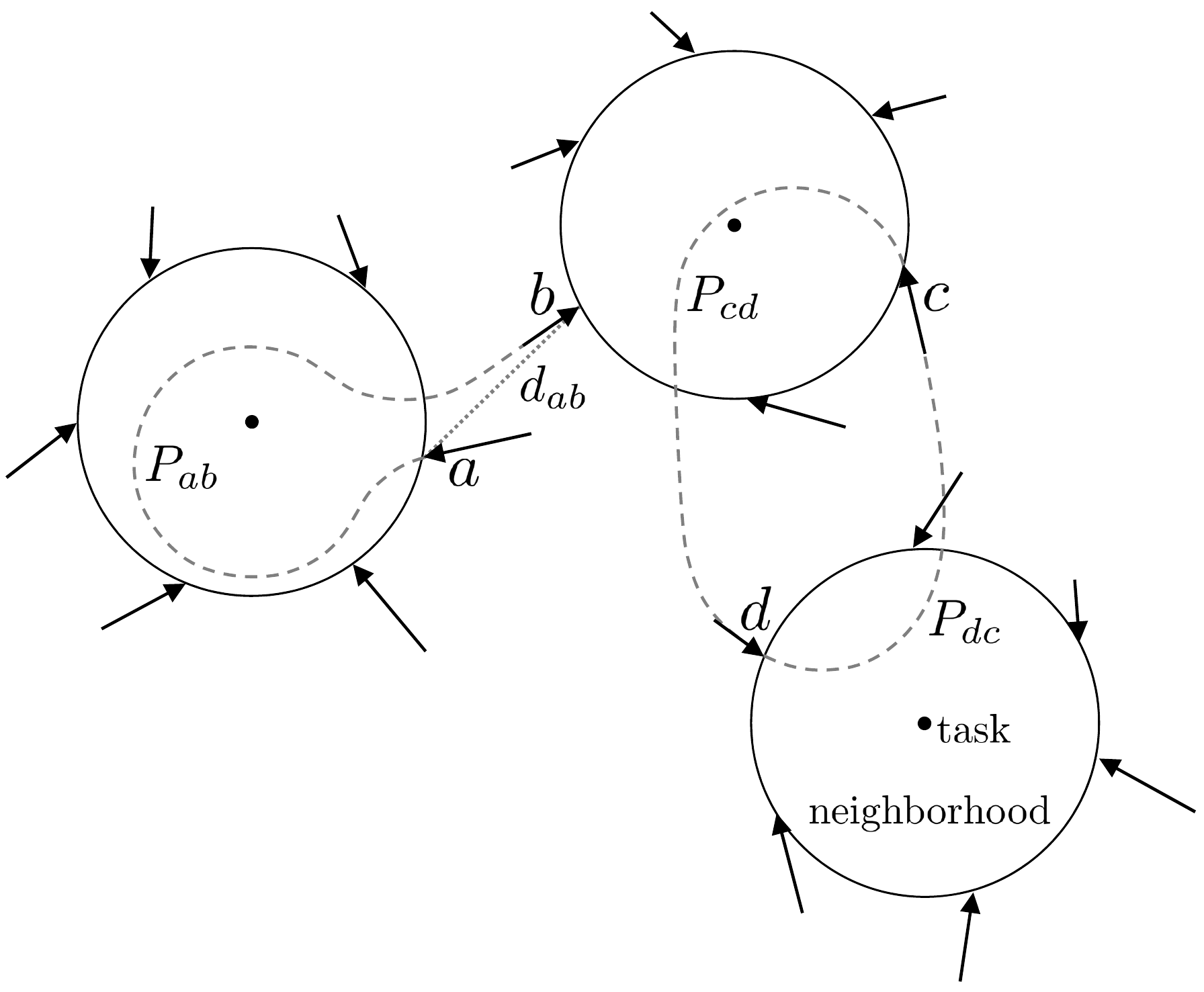}}
	\vspace*{0in}
	\caption{The state samples of a vehicle generated on the boundaries of given tasks' neighborhoods: the coordinates of each sample is pointed by the head of a vector, of which direction and magnitude signify the velocity of the sample. Two-way optimal paths are calculated for each pair of samples (e.g. $P_{cd}$ and $P_{dc}$) except some cases that the shortest Dubins paths are quite devious (e.g. $P_{ab}$).}
	\label{fig1}
	\vspace*{0in}
\end{figure}

For each pair of state samples of two different neighborhoods, the minimal time paths in both directions between the states are calculated by an optimal control solver (see Fig. \ref{fig1}).
To speed up the optimization process, the minimum Dubins path using the turning radius at the average speed of the two states and corresponding steering inputs are assigned as the initial guess; the speed of the UAV is assumed to be constant in this guess.
GPOPS-II \cite{GPOPS2} with SNOPT optimizer \cite{SNOPT} is used as the optimal control solver, and this numerical solver produces a local optimal path with state and control input histories.

The number of path calculations for constructing a complete roadmap quadratically increases with the number of samples per task and also with the number of tasks.
Since solving an optimal control problem is not a constant number of basic operations, but it is rather composed of time-consuming processes, the sampling-based roadmap construction might be the dominant procedure on the computation for the path planning with a nonlinear dynamic system, especially for practical UAV operations with scores of tasks.
Therefore, it is decided to reduce the number of path calculations by cutting-down some paths, according to a heuristic rule, which are hardly expected to be in the optimal solution of the planning.
It is conjectured that a devious and lengthy path between two samples compared with the straight-line distance is barely a part of the optimal tour (e.g. path $P_{ab}$ in Fig. \ref{fig1}).
Thus, at the phase of creating initial guess, if the shortest Dubins path using the minimum turning radius at $v_{\min}$ is longer than twice the straight-line distance between two samples' locations, then the numerical optimal path calculation is skipped and the path is not included in the roadmap.
The resultant reduced roadmap is deficient in a strict sense in that the resolution completeness in \cite{Obe12,Isa13} is no longer guaranteed.
However, the resolution completeness is only meaningful with a very large number of samples, which requires a massive computation, whereas the numerical simulations in section \ref{sec:sim} show that the refined tours from the reduced and the complete roadmaps of the same number of samples have little or no difference.

\subsection{Transformation to ATSP with Necessarily Intersecting Neighborhoods}\label{sec:alg_step2}

With the given sampled roadmap with optimal control paths, the path planning problem is approximately discretized as being GTSP, a variant of TSP, which is to find the shortest cycle visiting at least one node in every group.
Each sample in the roadmap corresponds to a {\it sample node} in the graph $G_R$ of the GTSP, and each path represents a directed edge of $G_R$ having the path cost (flight time) as its weight.
The samples generated along the boundary of the same task's neighborhood $N_i$ belong to the same group $V_i$, i.e. a subset of the vertex set $V(G_R)$ of the graph.
This graph can be transformed into ATSP using the Noon-Bean transformation and then be solved by exact or heuristic TSP solvers.
However, with a roadmap constructed from a finite number of samples, this approach might perform worse in the instances with intersecting neighborhoods of densely distributed tasks.

If a state sample $a$ on a neighborhood $N_a$ is also included in another neighborhood $N_b$, a vehicle passing this sample point obviously visits both of the neighborhoods.
Since $a$ is nominally generated to represent an entering state to $N_a$, a naive use of the roadmap searches a path that visits another sample generated for $N_b$ before or after the visit to $a$.
Therefore, even with a number of samples, it is likely that an unnecessarily circling path is induced in highly overlapping neighborhoods of dense tasks.
A modified transformation from GTSP to ATSP was proposed in \cite{Isa13}, based on the Noon-Bean transformation, to utilize the information on the visits at the samples that are included in multiple intersecting neighborhoods.
It was demonstrated that the proposed method in \cite{Isa13} outperforms or at least equals to the roadmap method using the Noon-Bean transformation.

\begin{figure}[t]
	\centerline{
		\includegraphics[width=.65\columnwidth]{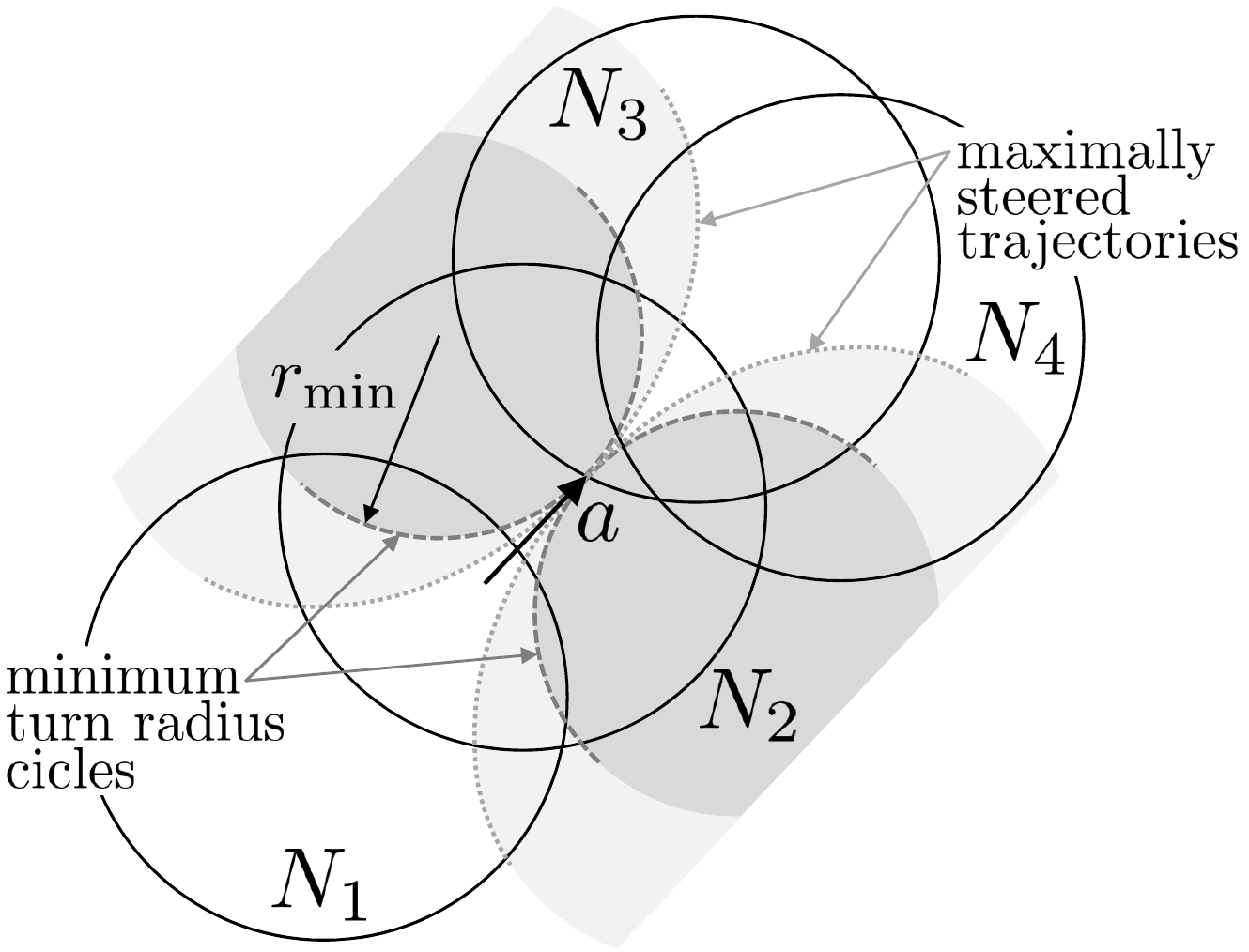}}
	\vspace*{0in}
	\caption{An example of necessarily intersecting neighborhoods of a UAV state sample. A UAV passing a state sample $a$ on the boundary of $N_3$ must visit the neighborhoods that intersect with both the left and right maximally steered trajectories before and after $a$, i.e. $N_1$, $N_2$, and $N_4$.}
	\label{fig2}
	\vspace*{0in}
\end{figure}

In this paper, the idea on the intersecting neighborhoods of samples is extended to extract more information about unavoidable visits of paths from sampled points.
Since UAV dynamics is non-holonomic, curvature-constrained, any local paths propagated before and after the state of a sample must be between the maximally steered trajectories in left- and right-wise directions (see Fig. \ref{fig2}).
Each maximally steered trajectory presents the local reachability of a vehicle, when it decelerates (accelerates) as far as possible with steering in one direction after the state (before the state).
If a neighborhood intersects with both of the left and right maximally steered trajectories, the UAV must visit the neighborhood before or after passing the sample, even in the case the neighborhood does not contain the sample.
The neighborhoods of this type are named as necessarily intersecting neighborhoods of the sample.
The necessarily intersecting neighborhoods of each sample can be partially obtained by simple calculations using the minimum turn radius circles at the slowest speed as in Fig. \ref{fig2} and are considered in the transformation presented in this paper.


Now, the original graph $G_R$ is transformed into an ATSP instance graph including {\it intersection nodes} representing necessarily intersecting neighborhoods.
The transformation is based on the Noon-Bean transformation and is functionally identical to the work in \cite{Isa13}, but it is presented in a concise but still accurate description.
The collection of the necessarily intersecting neighborhoods of an entry state sample $s$ is denoted as $\mathcal{N}_s\in\mathcal{N}$.
Then, for each state sample $s$ and corresponding sample node $v_s$, the following procedure is applied.

\begin{figure}[t]
	\centerline{
		\includegraphics[width=.65\columnwidth]{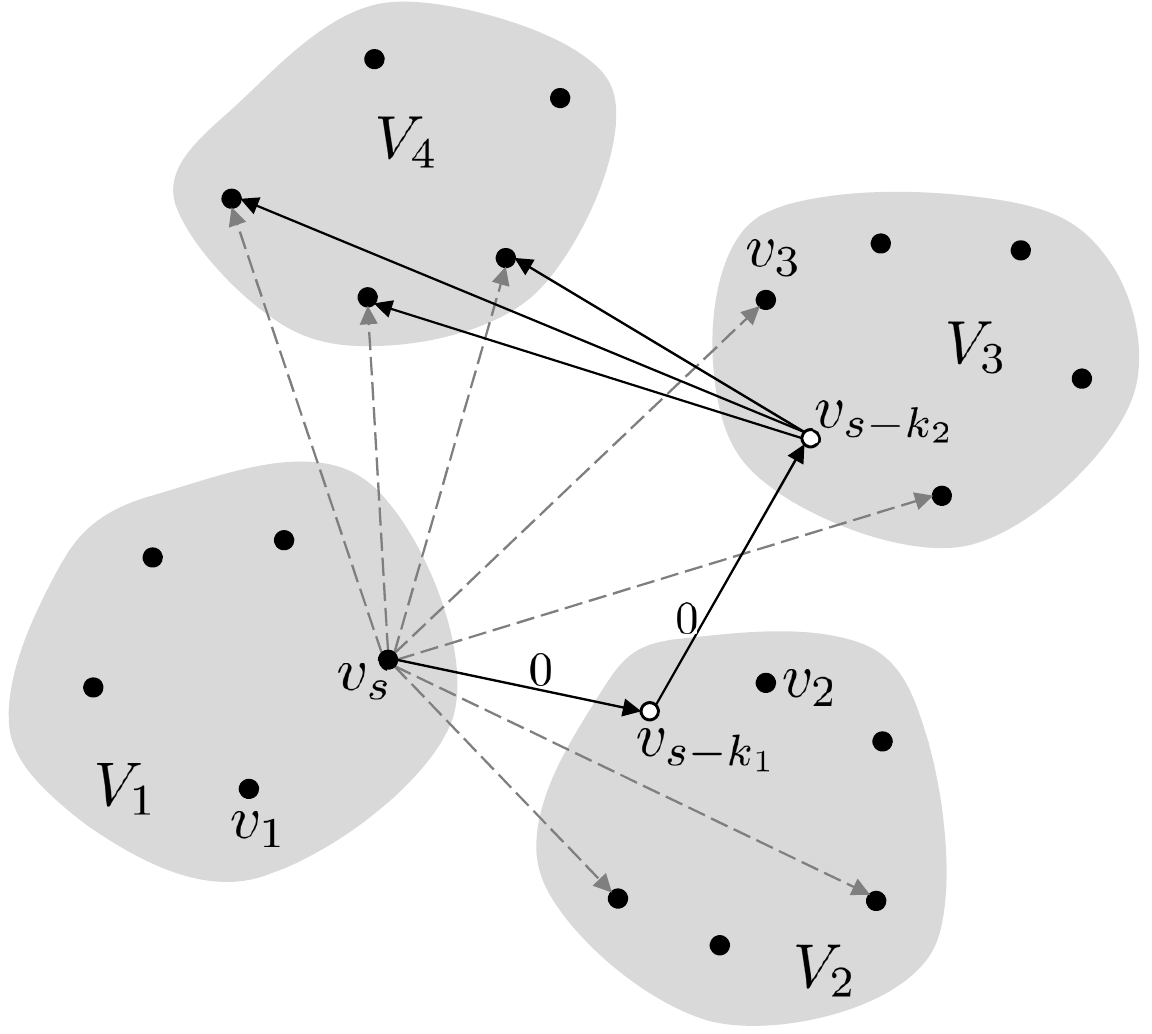}}
	\vspace*{0in}
	\caption{The first step on the graph transformation from the GTSP with necessarily intersecting neighborhoods to ATSP: The leaving edges from a sample node $v_s$ are transferred to an intersection node corresponding to a neighborhood in $\mathcal{N}_s$, and $v_s$ is connected with all of its intersection nodes via a zero-weight path.}
	\label{fig3}
	\vspace*{0in}
\end{figure}

If there is only one necessarily intersecting neighborhood $N_k$, i.e. $|\mathcal{N}_s|=1$, then,
\begin{enumerate}
	\item Add an intersection node $v_{s-k}$, corresponding to the intersection with $N_k$, to the graph as being affiliated to the node group $V_k$.
	\item Copy all leaving edges of the sample node $v_s$ to $v_{s-k}$ except the leaving edges to the nodes in $V_k$.
	\item Delete all leaving edges of $v_s$ and add an edge from $v_s$ to $v_{s-k}$ with a zero weight.
\end{enumerate}

Otherwise, if there are multiple $p$ necessarily intersecting neighborhoods, denoted as $N_{k_1},...,N_{k_p}$, and then,
\begin{enumerate}
	\item Add intersection nodes $v_{s-k_1},...,v_{s-k_p}$ corresponding to the intersections with $N_{k_1},...,N_{k_p}$, to the graph as being affiliated to $V_{k_1},...,V_{k_p}$, respectively.
	\item Copy all leaving edges of the sample node $v_s$ to $v_{s-k_p}$ except the leaving edges to the nodes in $V_{k_1},...,V_{k_p}$.
	\item Delete all leaving edges of $v_s$ and add an edge from $v_s$ to $v_{s-k_1}$ with a zero weight.
	\item Connect all the intersection nodes in sequence by a directed path from $v_{s-k_1}$ to $v_{s-k_p}$ with zero weights.
\end{enumerate}

The remaining process is the same as the Noon-Bean transformation:
\begin{enumerate}
	\item For each group, connect the nodes in the same group by an arbitrary cycle consisting of edges with zero weights.
	\item Add a large constant $M$ to the weights of all inter-group edges.
	\item For each node $v$, regardless of whether it is derived from a sample or an intersection, shift all the inter-group leaving edges of $v$ to the preceding node in the zero-weight cycle of its group: the tails of inter-group edges in the same group are cyclically shifted along the zero-weighted cycle in the reverse direction.
\end{enumerate}
It is sufficient that the large constant $M$ is selected to be larger than the sum of the $n$ largest weights in the graph \cite{Hel15}.

\begin{figure}[t]
	\centerline{
		\includegraphics[width=.65\columnwidth]{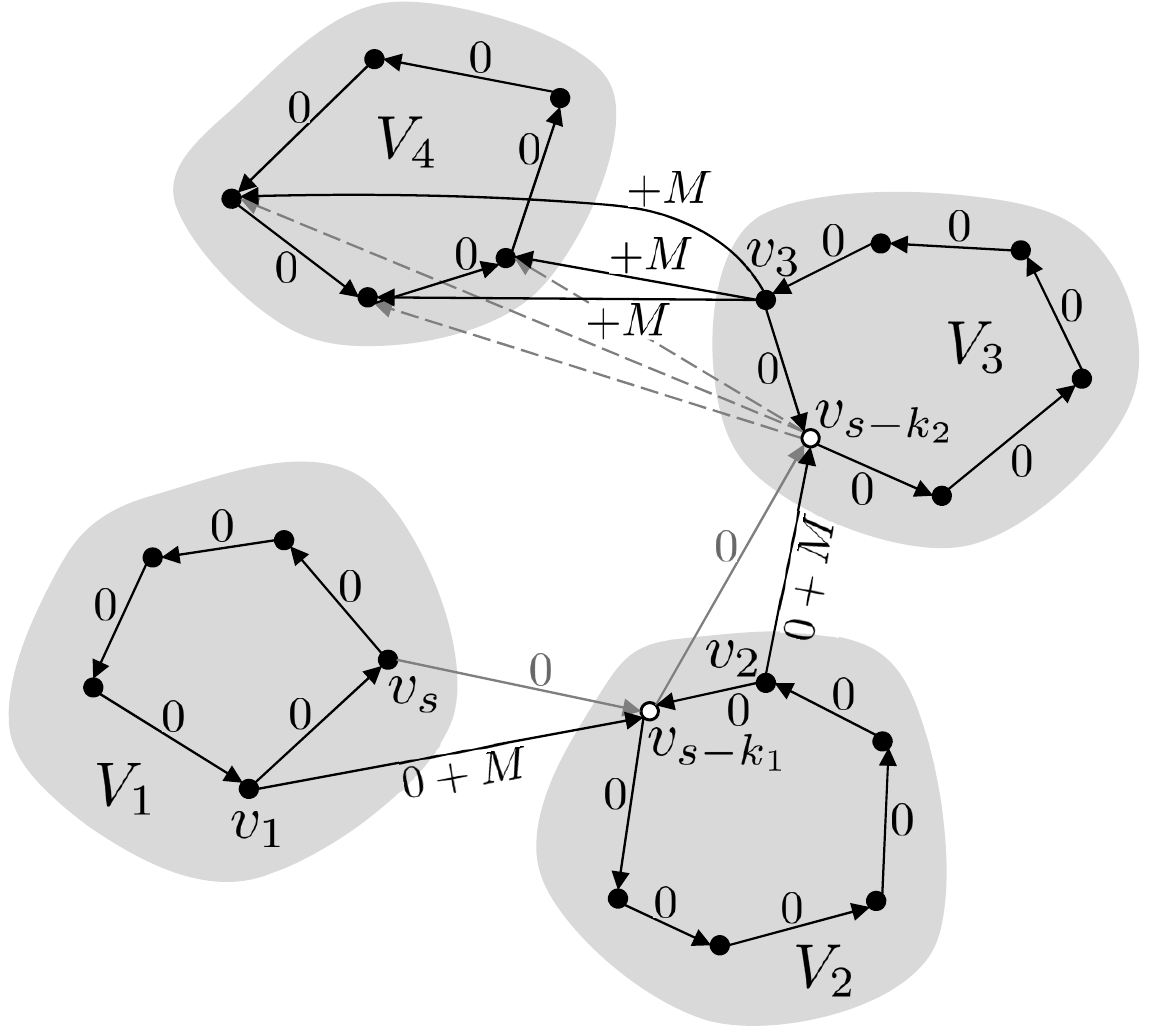}}
	\vspace*{0in}
	\caption{The second step on the graph transformation from the GTSP with necessarily intersecting neighborhoods to ATSP: the nodes in each group are linked by a zero-weight cycle; the inter-group leaving edges are cyclically shifted and their weights are increased by a large constant $M$.}
	\label{fig4}
	\vspace*{0in}
\end{figure}

A GTSP solution normally consists of $n$ nodes, each of which is one of the nodes in each group, whereas the transformed ATSP solution visits all given nodes, both from samples and intersections.
Since the nodes in a group are visited successively in the optimal solution of the ATSP, as an interpretation of the ATSP solution, the first visited node in a group is the visited node in the equivalent GTSP solution.
However, the original GTSP for the path planning does not include any intersection nodes, and thus the solution from the ATSP solver should be handled with some care as in the next subsection.

\subsection{Solving ATSP}\label{sec:alg_step3}

The next process is solving the transformed ATSP instance by a TSP solver.\footnote{Generic TSP solvers provide the transformation of ATSP to TSP.}
For a fast computation generating the order of visits, a well-implemented state-of-the-art heuristic solver, LKH-2 \cite{Hel09} is selected in this paper. 
LKH-2 is based on a revised implementation of the Lin-Kernighan heuristic \cite{Lin73}, a famous variable-depth local search algorithm for TSP.
The Lin-Kernighan heuristic searches the neighborhoods of a feasible solution derived by $k$-exchanges to find a $k$-opt solution, i.e. a local optimum, while determining the searching depth $k$.
The practical effectiveness of the Lin-Kernighan heuristic and its variations has been consistently reported \cite{Hel00,Ahu02,Reg11}.

The output of the ATSP solver is a sequence of all sample and intersection nodes representing a directed closed tour, and the first visited node of each group constitutes the solution of the equivalent GTSP in the Noon-Bean transformation.
Among the first visited nodes in the sequence, only the sample nodes are in the original GTSP for the path planning, and the intersection nodes are dummies representing the neighborhoods of some tasks that are concomitantly and inevitably visited during the flight passing the states of the ``first visited" sample nodes.
Thus, the concatenated paths connecting the first visited sample nodes compose a minimized tour in the sampling-based roadmap visiting all the given task neighborhoods.

\subsection{Tour Refinement}\label{sec:alg_step4}

The final step of the presented algorithm is a refinement of the tour from the ATSP solver.
Since any tour from the sampling-based roadmap is a crudely discretized solution from a practically limited number of samples in a 4-dimensional state space, the paths from the ATSP solver sometimes might look inefficient and the performance is quite restricted.
The sparsity of the same-sized sampling increases with the dimension of the dynamic system, and a higher number of samples for an elaborate approximation results in a significant increase of computation time.
Thus, in this paper, paths composing the tour from the previous step are locally refined by optimal control processes, which preserve the order of visiting neighborhoods in the tour while improve the solution quality by searching in the continuous state space.
Although the improvement is limited since the visiting order is fixed, this technique reduces the required number of samples for obtaining solutions with favorable quality in a practical time.

\begin{table*}[t]
	\centering{
		\caption{Statistical results with the complete roadmap in solving example instances}\label{tab1}
		\vspace*{-0.08in}
		\begin{tabular}{cccccccc}
			\hline\hline
			$n$& $m$  & roadmap      & \multicolumn{5}{c}{flight time (s)} \\ \cline{4-8}
			& 		  & construction  & without IN$^{\text a}$ & \multicolumn{2}{c}{with IN$^{\text a}$ from \cite{Isa13}} & \multicolumn{2}{c}{with necessarily IN$^{\text a}$} \\ \cline{5-8}
			& 		  &  time (m)      &  				 & \multicolumn{2}{c}{refinement} & \multicolumn{2}{c}{refinement} \\
			& 		  &                   &               & before & after &  before & after \\ \hline
			10	&	10	&	49.2, 1.2	 &	143.1, 7.1	&	119.3, 9.9	&	108.3, 7.8	&	106.9, 10.1&	98.3, 8.2	\\
			10	&	15	&	112.6, 4.5	 &	131.7, 3.8	&	110.6, 11.3&	101.2, 9.0	&	99.7, 12.0	&	91.8, 10.0	\\
			10	&	20	&	277.7, 43.0 &	123.4, 6.6	&	100.9, 10.0&	89.8, 8.6	&	91.2, 10.0	&	82.2, 7.0  \\
			15	&	10	&	160.6, 1.9	 &	190.9, 9.7	&	133.9, 8.6	&	122.4, 7.8	&	122.3, 8.8	&	112.7, 7.5	\\
			20	&	10	&	290.5, 3.4	 &	232.8, 8.2	&	168.0, 8.3	&	149.0, 9.5	&	145.0, 6.1	&	131.6, 8.5	\\ \hline\hline
			\multicolumn{8}{l}{\footnotesize Each data is the average value followed by the standard deviation from 10 test instances.}\\[-0.04in]
			\multicolumn{8}{l}{\footnotesize $^{\text a}$IN: intersecting neighborhoods.}
		\end{tabular}
	}
\end{table*}

\begin{table*}[t]
	\centering{
		\caption{Statistical results with the reduced roadmap in solving example instances}\label{tab2}
		\vspace*{-0.08in}
		\begin{tabular}{cccccccc}
			\hline\hline
			$n$& $m$  & roadmap      & \multicolumn{5}{c}{flight time (s)} \\ \cline{4-8}
			& 		  & construction  & without IN$^{\text a}$ & \multicolumn{2}{c}{with IN$^{\text a}$ from \cite{Isa13}} & \multicolumn{2}{c}{with necessarily IN$^{\text a}$} \\ \cline{5-8}
			& 		  &  time (m)      &  				 & \multicolumn{2}{c}{refinement} & \multicolumn{2}{c}{refinement} \\
			& 		  &                   &               & before & after &  before & after \\ \hline
			10	&	10	&	21.5, 3.1   &	143.6, 7.3  &	121.3, 9.2  &	107.6, 7.6   &	104.7, 13.0 &	95.3, 11.8	\\
			10	&	15	&	49.4, 7.2   &	131.7, 3.8  &	111.8, 10.9 &	102.1, 	9.6  &	100.3, 12.0 &	92.7, 10.2	\\
			10	&	20	&	120.8, 20.5 &	125.2, 5.6 &	101.8, 10.2 &	90.3, 8.1   &	91.7, 10.7  &	 84.2, 9.5	\\
			15	&	10	&	69.7, 7.4   &	191.0, 10.2 &	134.2, 8.7  &	121.9, 7.7   &	123.1, 8.1  &	114.5, 6.7	\\
			20	&	10	&	121.5, 11.6 &	236.1, 9.1 &	166.8, 10.2 &	146.2, 12.3 &	147.7, 6.7 &	136.0, 8.6	\\ \hline\hline
			\multicolumn{8}{l}{\footnotesize Each data is the average value followed by the standard deviation from 10 test instances.}\\[-0.04in]
			\multicolumn{8}{l}{\footnotesize $^{\text a}$IN: intersecting neighborhoods.}
		\end{tabular}
	}
\end{table*}

The tour refinement step is processed as follows.
First, the tour from concatenating the paths between the first visited sample nodes of the ATSP solution is re-partitioned by entering points of all neighborhoods along the tour.
Since some of the neighborhoods do not have {\it entering sample states} among the first visited sample nodes, starting at an arbitrary point on the tour, new entering states, which re-partition the tour, are created along the tour every time when a neighborhood is newly encountered.
The resultant tour has exactly $n$ paths and $n$ entering states.
The local refinement is performed between three successive entering states: the two end states are fixed and the optimally controlled path that intersects with at least one point in the neighborhood of the middle entering state is calculated by the same optimal control solver used for constructing the roadmap.
This refinement is repeated in an alternating order of the re-partitioned paths: even-numbered states may first refined while keeping the others and then odd-numbered states are refined in the same way.

\section{Numerical Simulations}\label{sec:sim}

This section describes numerical simulations performed to validate the computational efficiency and an enhancement in the solution quality of the presented algorithm, in comparison to the previous sampling-based roadmap algorithms.
The algorithms were implemented in a MATLAB environment on a PC with Intel{\scriptsize $^{\textregistered}$} Core{\texttrademark} i5-4670 3.40GHz CPU and 8.00GB RAM.
The optimal paths between sample states were calculated by using the MATLAB interface of GPOPS-II \cite{GPOPS2} with SNOPT optimizer \cite{SNOPT}, and the ATSP instances were solved by LKH-2 \cite{Hel09}.

\begin{figure}[t]
	\centerline{
		\includegraphics[width=.70\columnwidth]{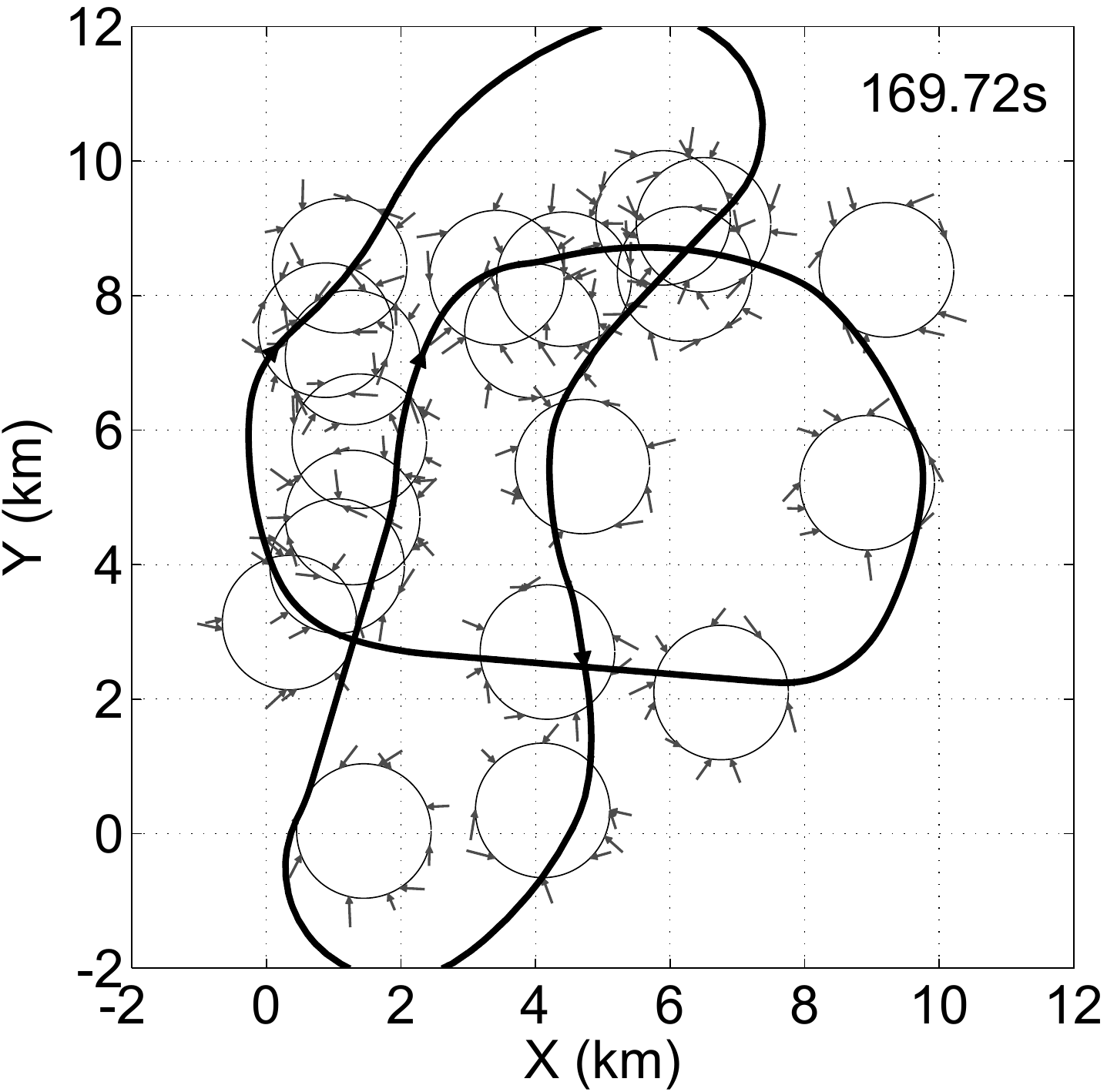}}
	\vspace*{0in}
	\caption{A solution tour for a problem with 20 tasks, derived from the sampling-based roadmap method using intersecting neighborhoods \cite{Isa13}. The complete roadmap is constructed from 10 samples per task.}
	\label{fig5}
	\vspace*{0in}
\end{figure}

For each of three problem sizes, i.e. $n=10, 15, 20$, ten test instances were created with randomly generated task points in a 10km$\times$10km plane, and each task had a circular neighborhood with a radius of 1km.
To identify the effect of the number of samples on the simulation results, three cases of $m=10, 15, 20$ were tested with the instances of 10 tasks, and 10 samples per task was used for other size problems.
The parameters in the UAV dynamics were assigned as follows: $v_{\min}=0.25$km/s, $v_{\max}=0.46$km/s, $c_1=0.1$, and $c_2=37.96m/s^2$, by setting $g=9.8m/s^2$ and $n=4$.
The minimum and maximum radii of coordinate turns under these parameters are 1.65km and 5.58km.

In this setting, ten different configurations of the sampling-based roadmap algorithms were tested, from the generic and complete roadmap method to the presented algorithm.
The configurations of algorithms are classified according to the roadmap consturction level (complete/reduced) and the use of intersection neighborhoods (none, intersection neighborhoods, necessarily intersecting neighborhoods), and the algorithms using intersection neighborhoods are sorted into the cases before and after the path refinement.

Table \ref{tab1} summarizes the simulation results from the algorithms using the complete roadmap of the same set of samples: all numeric data are given as averages and standard deviations, separated by commas, of the results from ten test instances.
The third column of Table \ref{tab1} lists the computation times for constructing the roadmaps in different cases of tasks and samples.
The times for other computations, including graph transformation, solving TSP, and path refinement, are below than 1\% of the roadmap construction times.
It is observed that the averaged solution qualities are improved as more information on the visits of neighborhoods from state samples are utilized, and this improvement becomes remarkable in the problems with 20 tasks, i.e. more densely distributed tasks.
The path refinement reduces the flight time by roughly around 10\% of the one before the refinement.
The increase of samples also contributes to the solution quality, but accompanied with a significant computational burden.
The simulation results of the same set of algorithms but using the reduced roadmap are presented in Table \ref{tab2}, and it is worth noting that comparable solutions can be obtained in less than a half of the computation time required for the algorithms using the complete roadmap.

\begin{figure}[t]
	\centerline{
		\includegraphics[width=.70\columnwidth]{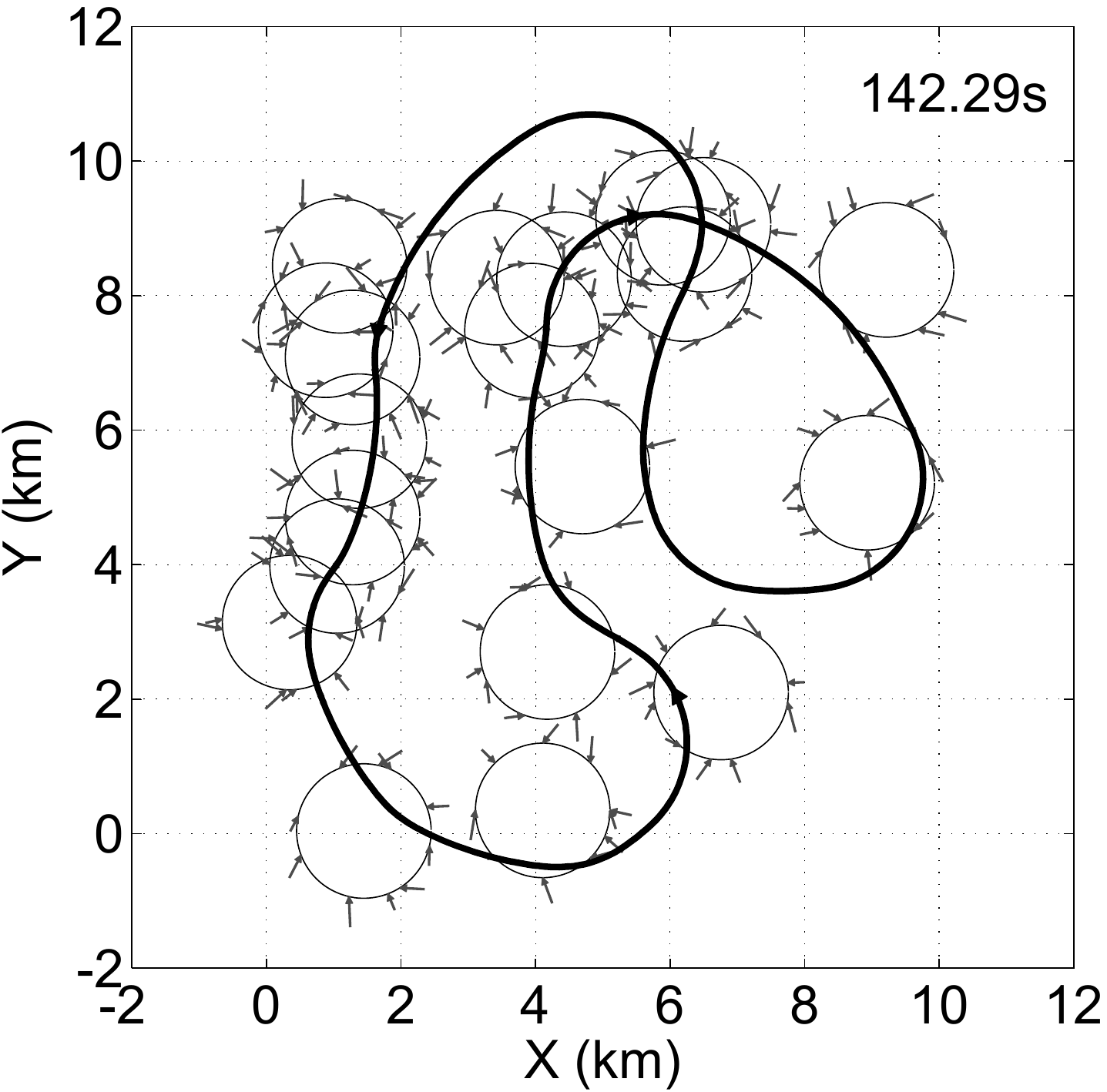}}
	\vspace*{0in}
	\caption{A pre-refined solution tour for a problem with 20 tasks, obtained by using necessarily intersecting neighborhoods. The complete roadmap is constructed from 10 samples per task.}
	\label{fig6}
	\vspace*{0in}
\end{figure}

\begin{figure}[t]
	\centerline{
		\includegraphics[width=.70\columnwidth]{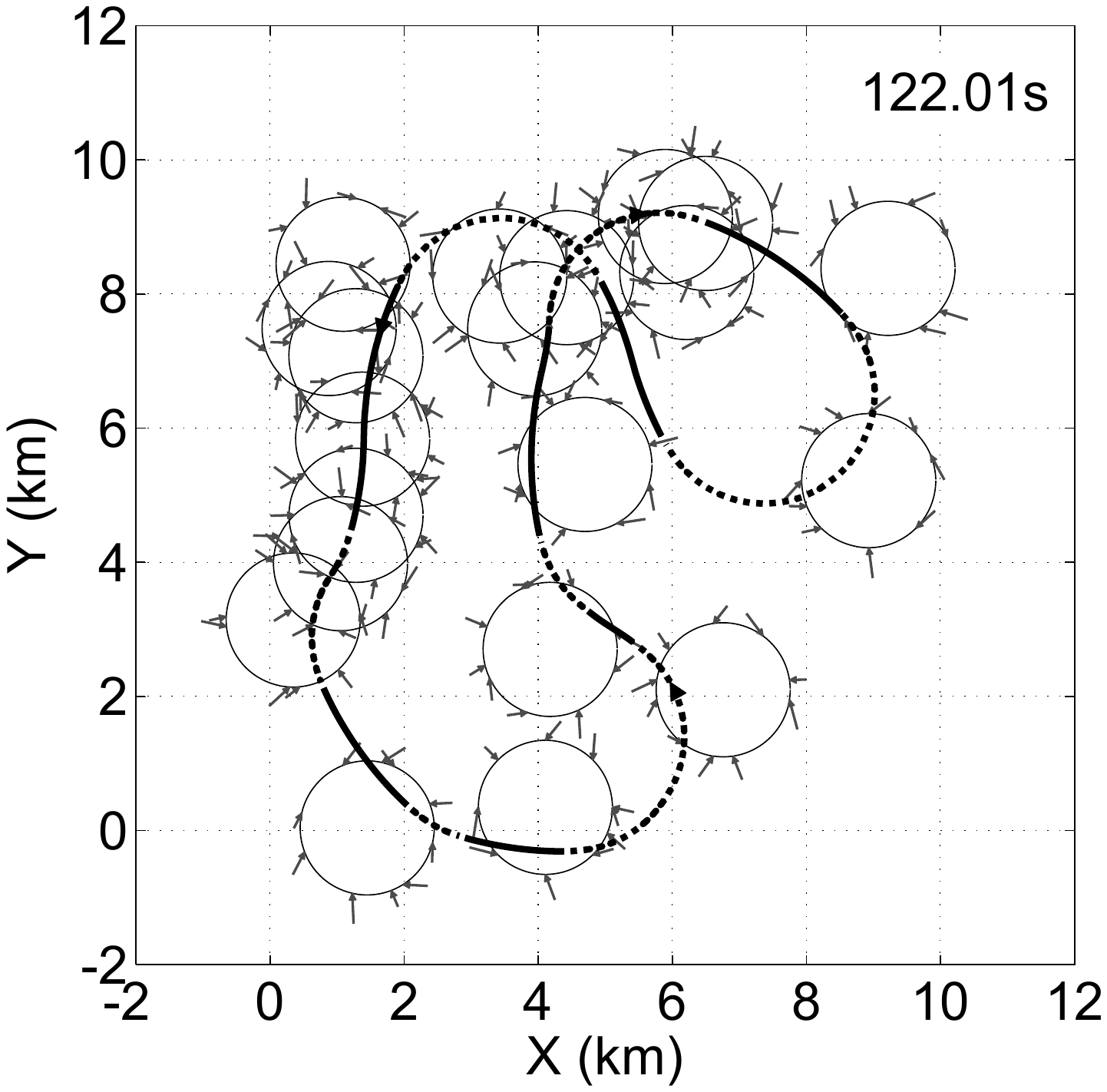}}
	\vspace*{0in}
	\caption{A refined solution tour for a problem with 20 tasks, obtained by using necessarily intersecting neighborhoods: high-speed sections in the tour are plotted as solid lines and low-speed sections are marked as dotted lines. The complete roadmap is constructed from 10 samples per task.}
	\label{fig7}
	\vspace*{0in}
\end{figure}

Figs. \ref{fig5} to \ref{fig7} depict the solution tours of different sampling-based roadmap algorithms for an example instance with 20 tasks.
The tours in Figs. \ref{fig5} and \ref{fig6} are pre-refined solutions from the algorithm using intersection neighborhoods in \cite{Isa13} and necessarily intersecting neighborhoods proposed in this paper.
Since the algorithm in \cite{Isa13} was devised to use the complete roadmap, all the results of the figures are based on the complete roadmap from the same number of samples: 10 samples per each of the tasks.
By extracting additional information on the intersections from the roadmap, the flight time in Fig. \ref{fig6} is shorten, comparing to the one in Fig. \ref{fig5}, and the flight tour appears more reasonable in terms of reducing deviated paths from neighborhoods.
The tour from the roadmap is locally improved by the refinement as in Fig. \ref{fig7}, which shows more tightly turning paths.
The tour in Fig. \ref{fig7} is divided into the parts moving slow (below $(v_{\min}+v_{\max})/2$, marked as dotted lines), and the other parts with high speed (above $(v_{\min}+v_{\max})/2$, marked as solid lines).
Fig. \ref{fig7} shows that a UAV following the locally optimized solution tour from the presented algorithm takes a low-speed for turning with a minimal radius and speeds up for nearly straight courses.



\section{Conclusions}

In this paper, a sampling-based roadmap algorithm using an optimal control approach has been presented to solve a dynamically-constrained TSPN, i.e. a path planning problem of a UAV performing remote sensing or wireless communication tasks.
With a minimal or no degradation of the final solution quality, the algorithm constructs a reduced roadmap with optimal control paths, excluding inefficient local paths, between state samples of the UAV.
For an improved solution quality with this reduced roadmap, the extra information of inevitable visits on neighborhoods of the non-holonomic UAV paths are extracted, and the optimized GTSP solution from the roadmap is locally refined in the continuous state space of the UAV.
Numerical simulations demonstrate that the presented algorithm produces improved solutions than the previous methods in a reduced computation time.



\section*{Acknowledgment}

This work was supported by Agency for Defense Development (contract:  \#UD140053JD).


\bibliographystyle{ieeetran}

\bibliography{acc16_TSPN}

\end{document}